\documentclass[letterpaper, 10 pt, conference]{ieeeconf}  

\IEEEoverridecommandlockouts                              

\usepackage{cite}
\usepackage{amsmath,amssymb,amsfonts}
\usepackage{algorithmic}
\usepackage{graphicx}
\usepackage{textcomp}
\usepackage{hyperref}
\usepackage{multirow}
\usepackage{xcolor}
\newtheorem{MyDef}{Definition}

\title{\LARGE \bf Graph and Recurrent Neural Network-based Vehicle Trajectory Prediction For Highway Driving}

\author{Xiaoyu Mo$^{1}$,
        Yang Xing$^{2}$,~\IEEEmembership{Member,~IEEE,}
        and~Chen~Lv$^{1}$,~\IEEEmembership{Senior Member,~IEEE}

\thanks{$^{1}$Xiaoyu Mo and Chen Lv are with the School of Mechanical and Aerospace Engineering, Nanyang Technological University, 639798, Singapore. (e-mail: xiaoyu006@e.ntu.edu.sg, lyuchen@ntu.edu.sg)}
\thanks{$^{2}$Yang Xing is with the Department of Computer Science, University of Oxford, OX1 3QD, UK. (e-mail: yang.xing@cs.ox.ac.uk)}
}

\begin{document}

\maketitle
\thispagestyle{empty}
\pagestyle{empty}

\begin{abstract}
Integrating trajectory prediction to the decision-making and planning modules of modular autonomous driving systems is expected to improve the safety and efficiency of self-driving vehicles. However, a vehicle's future trajectory prediction is a challenging task since it is affected by the social interactive behaviors of neighboring vehicles, and the number of neighboring vehicles can vary in different situations. This work proposes a GNN-RNN based Encoder-Decoder network for interaction-aware trajectory prediction, where vehicles' dynamics features are extracted from their historical tracks using RNN, and the inter-vehicular interaction is represented by a directed graph and encoded using a GNN. The parallelism of GNN implies the proposed method's potential to predict multi-vehicular trajectories simultaneously. Evaluation on the dataset extracted from the NGSIM US-101 dataset shows that the proposed model is able to predict a target vehicle's trajectory in situations with a variable number of surrounding vehicles. 
\end{abstract}

\section{INTRODUCTION}
Autonomous driving is expected to improve the safety and efficiency of our daily transportation thanks to the technological advancements in both algorithms and hardwares. While a typical autonomous driving system consists of four modules: perception, decision-making, planning, and control, researchers recently argue that autonomous vehicles will be safer if they can precisely predict future locations of its surrounding vehicles~\cite{li2019grip}. To this purpose, many trajectory prediction methods have been proposed, which fall in three categories, physics-based~\cite{ammoun2009real}, maneuver-based~\cite{hermes2009long, laugier2011probabilistic}, and interaction-aware methods~\cite{deo2018convolutional, li2019grip, mo2020recog, zhao2020tnt, mo2021heterogeneous}. More about this taxonomy can be found in~\cite{lefevre2014survey}. However, trajectory prediction is challenging in that driving is a complex interactive behavior~\cite{xing2021toward}, where the motion of a vehicle is affected by not only its driving style but also its surrounding vehicles, and the number of surrounding vehicles can be variant in different traffic situations.

Thanks to the availability of many real-world collected driving datasets~\cite{ushighway101, interactiondataset} and the success of neural networks, data-driven interaction-aware methods dominate the field of trajectory prediction in the last years. Most of these methods jointly consider temporal and spatial features~\cite{deo2018convolutional, zhao2019multi, mo2020recog}. Convolutional social pooling (CS-LSTM)~\cite{deo2018convolutional} applies long short-term memory network (LSTM)~\cite{hochreiter1997long} to individual vehicles' past tracks to extract their dynamics then aligns these dynamics into a target-centered occupancy grid to represent the spacial interaction. A CNN is then used to extract interaction feature from the grid. The performance of CS-LSTM can be affected by the size of the occupancy grid. It ignores the vehicle which is aggressively approaching the target vehicle but still outside the grid. Authors of~\cite{mo2020interaction} proposed to consider eight closet surrounding vehicles that have the most impact on the target vehicle's behavior rather than many vehicles in an occupancy grid. However, requiring the exact eight neighboring vehicles limited their model to be applied to situations where the number of surrounding vehicles varies.

Representing inter-vehicular interaction as a graph and applying graph neural network algorithms to model the interaction attracted great interest in the past two years~\cite{diehl2019graph, li2019grip, zhao2020tnt, mo2020recog}.
Authors of~\cite{diehl2019graph} conceptually proved that modeling a traffic scene as a graph to utilize the power of GNN increases prediction quality on a more-interactive highway dataset. They used only current information in their model and suggested integrating recurrent neural networks with GNNs in future works. GRIP~\cite{li2019grip} designed several graph convolutional blocks to extract interaction feature, which is then fed to an LSTM-based Encoder-Decoder to predict future trajectories. GRIP treats all the nodes equally when predicting a target vehicle's trajectory, which fails to emphasize the effects of the target vehicle's own dynamics. GRIP cannot accommodate state-of-the-art GNNs for interaction modeling to take advantage the advances in GNNs, like attention mechanisms. ReCoG~\cite{mo2020recog} modeled the relationships among vehicles and infrastructure as a heterogeneous graph and adopted state-of-the-art GNNs for the interaction feature. ReCoG focused on single vehicular trajectory prediction for urban driving, where the road structure affects vehicles' trajectories significantly. 

Inspired by~\cite{diehl2019graph}, this work improves the CNN-LSTM-based trajectory prediction method proposed in~\cite{mo2020interaction} by integrating RNNs and GNNs to handle the situation with varying number of surrounding vehicles and investigates the graph modeling's potential on the multi-vehicular trajectory prediction. The proposed model uses RNNs to extract dynamics features of all vehicles, then applies a GNN on a star-like directed graph, where a node corresponding to a vehicle contains its sequential feature and an edge from one node to another node implies that the latter's behavior is affected by the former, to summarize the inter-vehicular interaction. Finally, an RNN decoder is applied to the combination of the target vehicle's dynamics feature and its interaction feature for single vehicular trajectory prediction.

\begin{figure*}
    \centering
    \includegraphics[trim={0cm 0cm 0cm 0cm}, clip, width=1.0\textwidth]{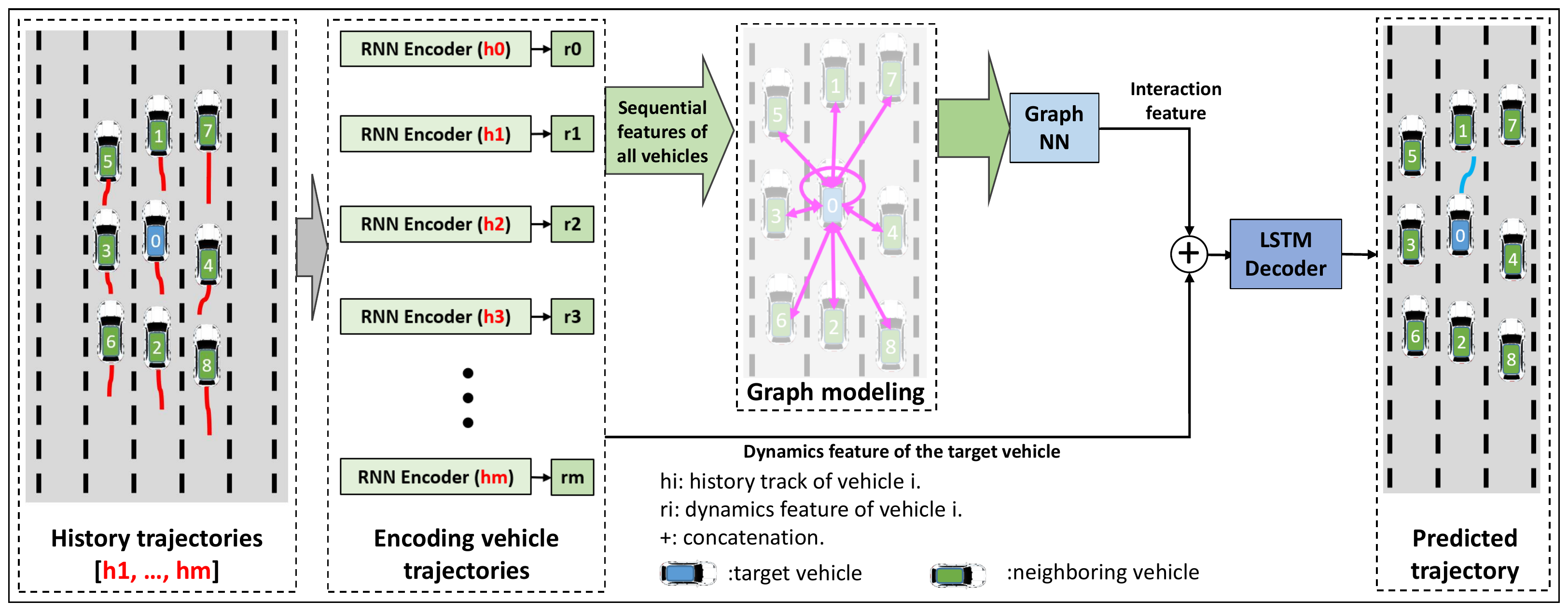}
    \vspace{-0.7cm}
    \caption{\textbf{Illustration of the proposed model in this study.} RNNs with shared weights are used to encode the dynamics features of vehicles individually. A GNN-based interaction encoder is applied to these dynamics features, which are contained in corresponding nodes in a directed interaction graph, to summarize the inter-vehicular interaction feature. Finally an LSTM decoder predicts the trajectory by jointly consider the target vehicle's dynamics and interaction features.}
    \label{fig: model}
    \vspace{-5mm}
\end{figure*}

The main contributions of this work can be summarized as follows:
\begin{itemize}
    \item A Graph-based interaction-aware trajectory prediction method is proposed.
    \item Ablative studies are conducted to show the necessity to jointly consider individual dynamics and interaction features.
    \item The potential of the proposed method to be applied to multi-vehicular trajectory prediction is investigated.
\end{itemize}

The rest of this paper is organized as below. Sec.~\ref{sec: method} expatiates the proposed method. Sec.~\ref{sec: experiment} describes experimental settings. Sec.~\ref{sec: results} evaluates the proposed model on the single trajectory prediction task and investigates its potential for multi-vehicular trajectory prediction. Finally, Sec.~\ref{sec: conclusion} concludes this paper and points out future directions.

\section{METHOD}
This section formulates the trajectory prediction problem and proposes a two-channel Encoder-Decoder structure, which consists of history encoder, interaction encoder, and future decoder, for this problem.
\label{sec: method}
\subsection{Problem Formulation}
\label{subsec: stpproblem}
This work aims to predict the future trajectory of a target vehicle driving on a highway given historical trajectories of its up-to-eight surrounding vehicles. As shown in Fig.~\ref{fig: model}, this task considers two kinds of vehicles: the target vehicle and its neighboring vehicles. 

Neighboring vehicles considered are the target vehicle's preceding (\#1) and following (\#2) vehicles, its nearest neighbors in adjacent lanes (\#3 and \#4), in terms of longitudinal distance, and their preceding (\#5 and \#7) and following (\#6 and \#8) vehicles.

The input to the model ($\mathcal{H}_t$) is a set of historical trajectories of all considered vehicles, including the target vehicle. 
\begin{equation}
    \mathcal{H}_t = \lbrace h^0_t, h^1_t, \cdots, h^{m}_t \rbrace
    \label{eq: Histstp}
\end{equation}
where $ h^i_t = [p^i_{t-T_h+1}, p^i_{t-T_h+2}, \cdots, p^i_t]$ represents the sequence of historical trajectory of vehicle $i$ at time $t$. $T_h$ is the traceback horizon. Without loss of generality, this work numbers the target vehicle as $0$ and the neighboring vehicles from $1$ to $m \in [1, 8]$.

The output is the predicted future trajectory of the target vehicle at time $t$:
\begin{equation}
    f^0_t = [p^i_{t+1}, p^i_{t+2}, \cdots, p^i_{t+T_f}]
    \label{eq: Futstp}
\end{equation}
where $T_f$ is the prediction horizon.

\subsection{Model Structure}
To solve the single trajectory prediction problem, this work proposes a GNN-RNN based model, which is designed under the Encoder-Decoder structure and consists of two encoders (history encoder, interaction encoder) and one decoder (future decoder). The history encoder, implemented with an RNN, extracts an individual vehicle's dynamics from its historical trajectory. The interaction encoder uses a GNN to summarize interaction features among a variable number of vehicles. Then the future decoder uses another RNN to roll out the future trajectory of the target vehicle. Details of these main parts of the proposed model are described below.
\subsubsection{History Encoder}
The history encoder is shared across all vehicles to encode individual dynamics from their own historical trajectories. Eq.~\ref{eq: histenc} shows that the encoder is applied to historical tracks of all vehicles in parallel.
\begin{equation}
    R_t = \{r^0_t, r^1_t, \cdots, r^m_t \} = \mathrm{RNN_{hist}}(\mathrm{Emb}(\mathcal{H}_t)), 
    \label{eq: histenc}
\end{equation}
where $\mathrm{Emb}()$ is a linear transformation embedding the low-dimensional  xy-coordinates into a high-dimensional vector space, $\mathrm{RNN_{hist}}$ is a shared RNN applied to the embedded historical tracks of all vehicles, $r^i_t$ is the dynamics feature of vehicle $i$ at time $t$.

\subsubsection{Interaction Encoder}
\label{subsec: intenc}
Considering the fact that driving is an interactive activity and the mutual influence between two cars on each other is different, this method models the inter-vehicular interaction as a directed graph, where each node represents a vehicle and contains the vehicle's sequential feature. 
\begin{MyDef}[Directed Graph]
A graph can be represented by $\mathbb{G} = (V, E)$, where $V=\{v_0, \cdots, v_m\}$ is the set of $m+1$ nodes, and $E \subset V \times V$ is the set of edges.
If the edge from node $i$ to node $j$ is different from the edge from node $j$ to node $i$, the graph is a directed graph. 
\end{MyDef}

Since this work models the interaction among vehicles as a graph, the structure of the graph will significantly affect the performance and efficiency of method~\cite{diehl2019graph}. If the graph contains only self connections, its performance should be similar to a simple model working on the target vehicle's historical track only. While if the graph contains all connections (every node is connected to the rest of the nodes), it considers redundant connections, which increases quadratically with the number of nodes.  This work considers up-to-eight neighboring vehicles and constructs the interactive graph as a star-like graph.

\textbf{Graph Construction.} Without loss of generality, this work sets the target vehicle as $v_0$, and all the neighboring vehicles as $\{ v_1, \dots, v_m \}$. Then the edge set of the star-like graph with self-loop is constructed.
\begin{equation}
    E = \{ e^{0, j} \}_{(j=0, \cdots, m)} \cup \{ e^{j, 0} \}_{(j=1, \cdots, m)},
\end{equation}
where $e^{i,j}$ means that there is a directed edge from node $j$ to node $i$, that is, node $j$ is the neighbor of node $i$ and node $j$'s behavior will affect node $i$'s behavior. An example of the star-like directed graph with self-loop can be found in Fig.~\ref{fig: model}

Nodes in the constructed graph contain corresponding vehicles' sequential features $r^i_t$ and directed edges represent their directed effects to others. Then the graph is processed by a graph neural network to model the the interaction feature $g^0_t \in G_t$ as shown in Eq.~\ref{eq: interenc}:
\begin{equation}
    G_t = \mathrm{GNN_{inter}}(R_t, E_t),
    \label{eq: interenc}
\end{equation}
where $E_t$ represents the graph structure at time $t$, $\mathrm{GNN_{inter}}$ is the interaction encoder implemented with a 2-layer GNN, and $G_t=\{ g^0_t, \cdots, g^m_t \}$ contains the interaction features of all vehicles at time $t$.

\subsubsection{Future Decoder}
The future trajectory $f^0_t$ is predicted upon the target vehicle's dynamics feature $r^0_t$ and interaction feature $g^0_t$ using another RNN.
\begin{equation}
    f^0_t = \mathrm{RNN_{fut}} ([g^0_t, r^0_t]), 
\end{equation}
where $\mathrm{RNN_{fut}}$ is the future decoder implemented with RNN and $[g^0_t, r^0_t]$ is the concatenation of $g^0_t$ and $r^0_t$. 

The model also uses proper fully-connected layers, which are not shown in the equations. Further details can be found in Sub.Sec.~\ref{subsec: implement} and the released code.

\section{EXPERIMENTAL SETUP}
\label{sec: experiment}
The experiments are set up with data pre-processing, model implementing, and metric setting.
\subsection{Dataset}
This work uses vehicle trajectories extracted from the publicly available NGSIM US-101~\cite{ushighway101} dataset, collected from 7:50 a.m. to 8:35 a.m. on June 15, 2005, for training and validation. The study area is a 640 meters segment of U.S. Highway 101, consisting of five main lanes, one auxiliary lane,  and on-ramp and off-ramp lanes. The vehicle trajectory data are recorded at 10 Hz using eight synchronized digital video cameras mounted from the top of a 36-story building. This work selects roughly balanced data so that the lane-keeping trajectories do not dominate the dataset.

\subsection{Data Pre-processing}
This work first selects target vehicles then selects data pieces from their trajectory.
\subsubsection{Target Vehicles Selection} A vehicle is selected as a target vehicle upon following conditions:
\begin{itemize}
    \item It has not been driven in lanes 7 (On-ramp) and 8 (off-ramp).
    \item It only changed its lane once during the recording time.
    \item Its recorded track is at least 1,000 feet in length.
    \item Its lane-change maneuver happened within the range from 300 to 1,900 feet in the study area.
    \item Its lane-change maneuver was obvious that the maximum lateral displacement before and after lane-change is greater than 10 feet.
\end{itemize}

This step finally selects 124 ( out of 1,993) vehicles from the \textit{07:50am-08:05am} segment, 106 (out of 1,533) vehicles from the \textit{08:05am-08:20am} segment, and 68 (out of 1,298) vehicles from the \textit{08:20am-08:35am} segment. 

\subsubsection{Data Selection}
For a target vehicle, 260 frames from 13 seconds (130 frames) before lane-change to 13 seconds (130 frames) after lane-change are considered as candidates of the current frame (time $t$ in Eq.~\ref{eq: Histstp}). Then a data is stored in the dataset if the following conditions are all satisfied:
\begin{itemize}
    \item The target vehicle has a 3-second historical trajectory and a 5-second future trajectory.
    \item All neighboring vehicles have a 3-second historical trajectory.
\end{itemize}

This step selects totally 63,176 pieces of data with 23,803 from the \textit{07:50am-08:05am} segment, 24,559 from the \textit{08:05am-08:20am} segment, and 14,814 from the \textit{08:20am-08:35am} segment.

\textbf{Translation.}
A stationary frame of reference with its origin fixed at the target vehicle's current position is used for each data piece. 

\textbf{Down-sampling.}
The raw data in NGSIM US-101 is recorded with a sampling rate of 10 Hz. This work down-sample the historical tracks by a factor of 2 and the future trajectories by 5.

\textbf{Edge indexes.} 
The edge set representing the graph structure is constructed as described in SubSec.~\ref{subsec: intenc}.

\textbf{Data format.}
A data with 3 parts is stored to the dataset. 
\begin{equation}
    data = \{ \mathcal{H}_t, E_t, y_t \},
\end{equation}
where $\mathcal{H}_t$ is the historical tracks of all vehicles, $E_t$ is the edge set containing the structure of the interactive graph, and $y_t$ is the target vehicle's ground truth future trajectory.

After the above processing, this work randomly selects 10,000 data pieces from the whole dataset as the validation set and uses the rest of the dataset for training.

\subsection{Implementation Details}
\label{subsec: implement}
All the models in this work are implemented with PyTorch~\cite{NEURIPS2019_9015} except the GNN layers, which are implemented with PyTorch Geometric~\cite{Fey/Lenssen/2019}. The history encoder is implemented using a one-layer Gated Recurrent Unit (GRU)~\cite{chung2014empirical} with a 32-dimensional hidden state, and the future decoder is implemented using a two-layer LSTM with a 64-dimensional hidden state. The interaction encoder is implemented with two Graph Attention Network (GAT)~\cite{velivckovic2017graph} layers, which adopt concatenated three-head attention mechanism to stabilize the training process. This work uses LeakyReLU with a 0.1 negative slope as the only activation function. 

The proposed model is trained for 50 epochs to minimize the same loss function as described in~\cite{mo2020interaction} using Adam~\cite{kingma2014adam} with a learning rate of 0.001. Full implementation of the proposed model can be found in the released code.

\subsection{Metrics}
This work uses root-mean-square error (RMSE) in meters of the predicted trajectories against the ground truth future trajectories to evaluate different models. RMSE is calculated for each predictive time step $t_p$ within 5 seconds in the future. Previous works~\cite{mo2020interaction, deo2018convolutional, jeon2020scale} also adopt this metric. 
\begin{equation}
    RMSE(t_p) = \sqrt{\frac{1}{n}\sum^{n}_{i=1}( (\hat{x}^i_{t_p} - x^i_{t_p} )^2 + (\hat{y}^i_{t_p} - y^i_{t_p} )^2 )},
    \label{eq: metric}
\end{equation}
where $n=10000$ is the size of test set, $(\hat{x}^i_{t_p}, \hat{y}^i_{t_p})$ and $(x^i_{t_p}, y^i_{t_p})$ are the predicted position of the target vehicle in data $i$ at time ${t_p}$ and the corresponding ground truth, respectively.

\section{RESULTS AND DISCUSSION}
\label{sec: results}
This section compares the proposed two-channel model with its ablations and previous works on the single trajectory prediction task, followed by an investigation of its potential for multiple trajectory prediction.

\subsection{Single Trajectory Prediction (STP)}
Following methods are implemented for comparison:
\begin{itemize}
    \item \textbf{Dynamics-only}: this is the one-channel ablation of the proposed model considering the target vehicle's dynamics feature only for prediction.
    \item \textbf{Interaction-only}: this is the other one-channel ablation using the interaction feature extracted by the GNN only.
    \item \textbf{Two-channel}: this is the proposed two-channel model.
\end{itemize}
The above implementations are trained and validated using the same dataset.

Results reported in some related works are also listed in Tab.~\ref{tab: othersresults}. However, this work focuses on comparing results between the proposed method and its ablations, considering that different works are using different training and validation datasets.

\begin{figure}
    \centering
    \includegraphics[trim={0cm 0cm 0cm 0cm}, clip, width=1.0\columnwidth]{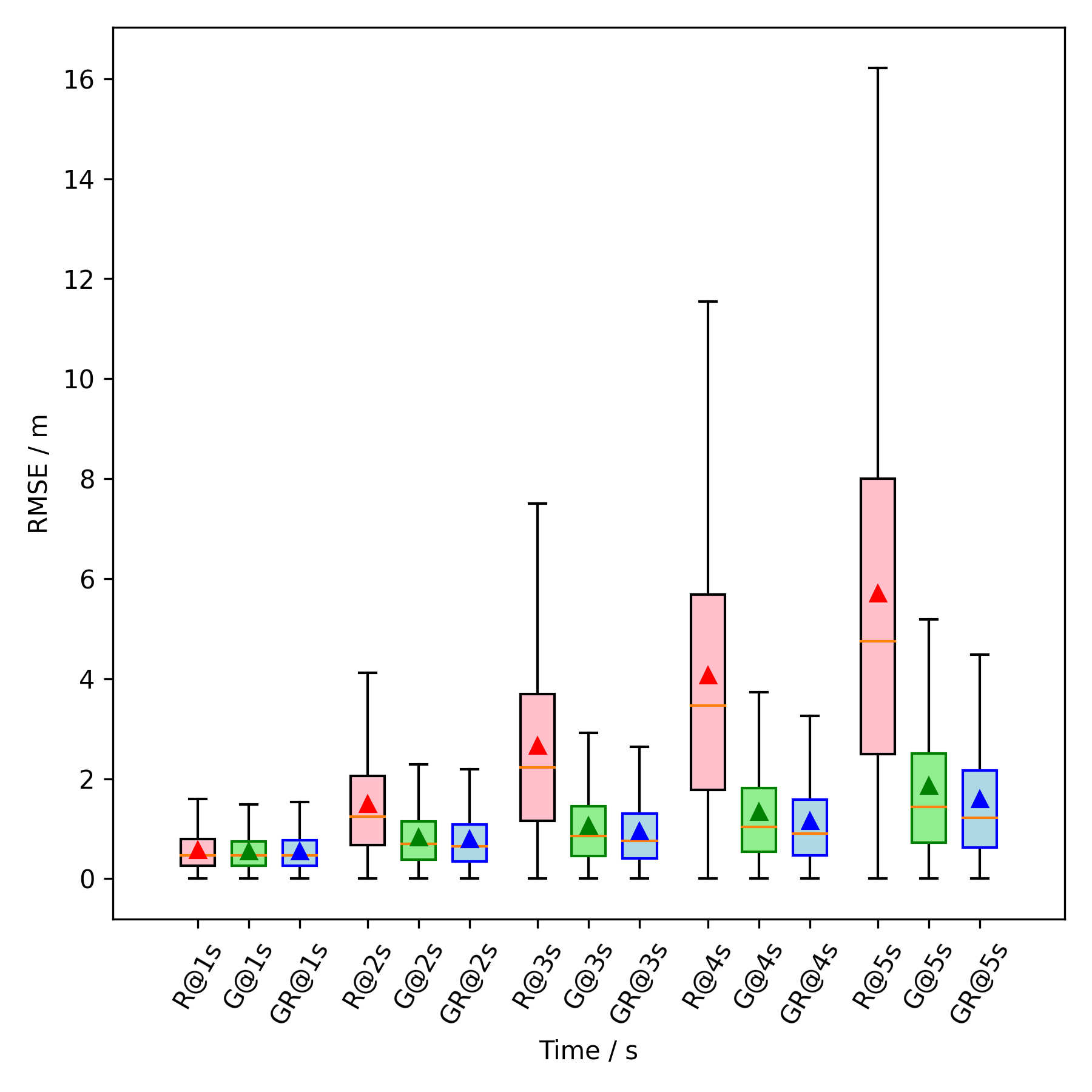}
    \vspace{-7mm}
    \caption{\textbf{Box plots of the RMSE of implemented models.} R is the dynamics-only model, G the interaction-only model, and GR the proposed two-channel model.}
    \label{fig: stpbox}
    \vspace{-5mm}
\end{figure}

\begin{table}[!ht]
\caption{\textbf{Prediction performance comparison (RMSE in meters)}}
\centering
\begin{tabular}{|c|c|c c c c c|} 
\hline
&\multirow{2}{6em}{ \textbf{Methods }} &\multicolumn{5}{c|}{\textbf{Prediction horizon}}\\
\cline{3-7} 
& & \textbf{\textit{1 sec}}& \textbf{\textit{2 sec}}& \textbf{\textit{3 sec}} & \textbf{\textit{4 sec}} & \textbf{\textit{5 sec}} \\
\hline
1 & Dynamics-only (Ours) & 0.74 & 1.86 & 3.30 & 5.07 & 7.11 \\
\hline
2 & Interaction-only (Ours) & 0.67 & 1.03 & 1.34 & 1.74 & 2.46 \\
\hline
3 & Two-channel (Ours) & 0.68 & 0.99 & 1.21 & 1.53 & 2.14 \\
\hline
\hline
4 & CS-LSTM~\cite{deo2018convolutional} & 0.61 & 1.27 & 2.09 & 3.10 & 4.37 \\
\hline
5 & GRIP~\cite{li2019grip} & 0.37 & 0.86 & 1.45 & 2.21 & 3.16 \\
\hline
6 & CNN-LSTM~\cite{mo2020interaction} & 0.64 & 0.96 & 1.22 & 1.53 & 2.09 \\
\hline
\end{tabular}
\vspace{0.2cm}
\label{tab: othersresults}
\end{table}
Tab.~\ref{tab: othersresults} compares different models. It shows that:
\begin{itemize}
    \item Interaction-aware methods (2,3,4,5,6) outperform the dynamics-only method (1). This demonstrates the necessity of modeling interactions for trajectory prediction as stated in previous works~\cite{deo2018convolutional, mo2020interaction}.
    \item The proposed two-channel model outperforms its interaction-only ablation. This shows that the target vehicle's dynamics feature should be emphasized in some way for trajectory prediction. This work sets an additional channel for it.
    \item The proposed method matches the CNN-LSTM method with advances in considering variable number of surrounding agents and the potential for multi-trajectory prediction.
    \item The proposed method outperforms GRIP and CS-LSTM in longer-term prediction (3-5sec). However, for the short-term prediction, GRIP shows better performance possibly in that GRIP uses the whole dataset from NGSIM, where the lane-keeping trajectories are dominant and less challenging for trajectory prediction.
\end{itemize}

Fig.~\ref{fig: stpbox} shows box plots of the RMSE errors of models implemented in this study over a 5-second time in the future, where the red boxes are the results of the dynamics-only model (R), the green boxes the results of the interaction-model (G), and blue boxes the proposed two-channel model (GR). Triangles in a box represents its mean value. Outliers are ignored for clarity. In addition to Tab.~\ref{tab: othersresults}, Fig.~\ref{fig: stpbox} shows that the prediction of interaction-aware methods (G \& GR) is more stable (shorter interquartile range (IQR)) than dynamics-only model (R) and the proposed two-channel model produces the shortest IQR. Please note that the mean value shown in Fig.~\ref{fig: stpbox} is calculated using Eq.~\ref{eq: boxmean}:
\begin{equation}
    RMSE(t_p) = \frac{1}{n}\sum^{n}_{i=1} \sqrt{(\hat{x}^i_{t_p} - x^i_{t_p} )^2 + (\hat{y}^i_{t_p} - y^i_{t_p} )^2 },
    \label{eq: boxmean}
\end{equation}
which is slightly different to the results in Tab.~\ref{tab: othersresults}.

\begin{figure}
    \centering
    \includegraphics[trim={0cm 0cm 0cm 0cm}, clip, width=1.0\columnwidth]{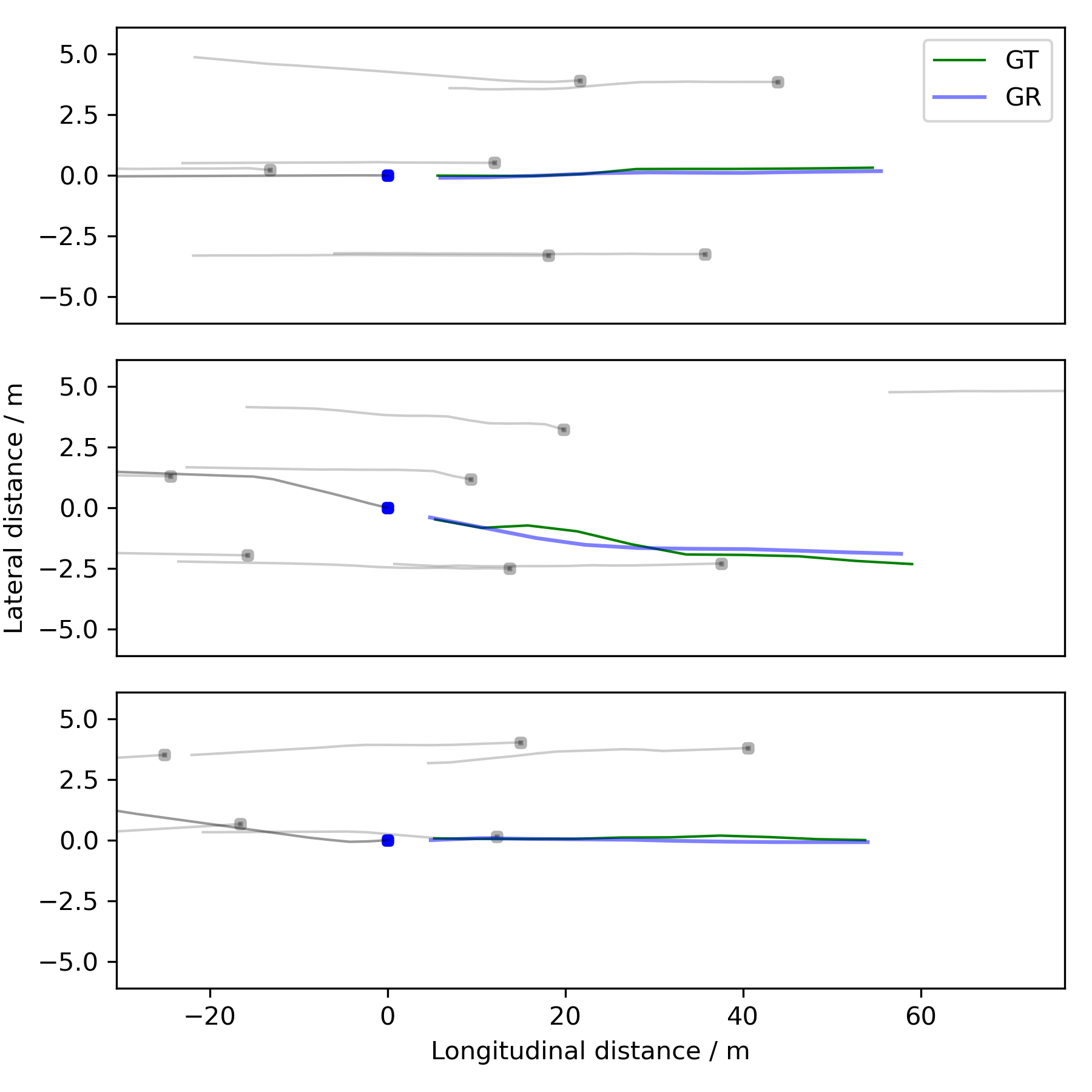}
    \vspace{-7mm}
    \caption{\textbf{Visualized STP predictions.} Squares are the considered vehicles (target vehicle in blue and neighboring vehicles in gray). Gray lines are the vehicles' historical tracks over the last 3 seconds. The green line is the ground truth (GT) future trajectory of the target vehicle. The blue line is the prediction of the proposed two-channel model (GR). All the vehicles move from left to right.}
    \label{fig: stpvis}
    \vspace{-5mm}
\end{figure}

Fig.~\ref{fig: stpvis} visualizes prediction results in situations with different numbers of surrounding vehicles from the validation set. It shows that the proposed model can predict the target vehicle is going to keep or change lane in the next 5 seconds regardless of how many surrounding vehicles are in sight.

Even though this work focuses on single trajectory prediction, the proposed model has the potential to be applied to multi-vehicular trajectory prediction since the interaction encoder implemented with GNN processes all nodes simultaneously, see Eq.~\ref{eq: interenc}. The following section briefly formulates the problem of multi-vehicular trajectory prediction (MTP) and shows the proposed method's performance on MTP.

\subsection{Multiple Trajectory Prediction (MTP)}
From the ego vehicle's point of view, MTP wants to predict future trajectories of up-to-eight target vehicles based on historical tracks of more vehicles. In this formulation, considered vehicles are separated into three categories: one ego vehicle, up-to-eight target vehicles, and some other surrounding vehicles. The MTP problem here is formulated similar to Sub.Sec.~\ref{subsec: stpproblem} and the target vehicles are selected as the selection of neighboring vehicle in Sub.Sec.~\ref{subsec: stpproblem}. Please note that this part is only to investigate the proposed method's potential for multi-agent setting, and the only difference to STP is the input and output data.

The input to the model is historical trajectories of all considered vehicles, 
\begin{equation}
    \mathcal{H}_t = \lbrace h^0_t, h^1_t, \cdots, h^{m}_t, h^{m+1}_t, \cdots, h^n_t \rbrace,
    \label{eq: Hist}
\end{equation}
where the $h^0_t$ is the ego vehicle's historical track and $1\leq m \leq 8$ is the number of target vehicles. MTP simultaneously predicts $m$ target vehicles' future trajectories, numbered from $1$ to $m$, based on historical trajectories of $n+1$ vehicles.

The output is then the predicted future trajectories of the target vehicles:
\begin{equation}
    \mathcal{F}_t = \lbrace f^1_t, f^2_t,\cdots, f^{m}_t \rbrace,
    \label{eq: Fut}
\end{equation}
where $ f^i_t = [p^i_{t+1}, p^i_{t+2}, \cdots, p^i_{t+T_f}]$ represents the sequence of future trajectory of vehicle $i$ at time $t$.

The dataset used here is pre-processed from the \textit{08:05am-08:20am} segment of NGSIM US-101. The size of training and validation datasets are 533,564 and 13,3392, respectively. 
 
\begin{table}[!ht]
\caption{\textbf{MTP performance comparison (RMSE in meters)}}
\centering
\begin{tabular}{|c|c|c c c c c|} 
\hline
&\multirow{2}{6em}{ \textbf{Methods }} &\multicolumn{5}{c|}{\textbf{Prediction horizon}}\\
\cline{3-7} 
& & \textbf{\textit{1 sec}}& \textbf{\textit{2 sec}}& \textbf{\textit{3 sec}} & \textbf{\textit{4 sec}} & \textbf{\textit{5 sec}} \\
\hline
1 & Two-channel (Ours) & 0.54 & 1.12 & 1.80 & 2.63 & 3.67 \\
\hline
\hline
2 & GRIP(ALL)~\cite{li2019grip} & 0.64 & 1.13 & 1.80 & 2.62 & 3.60 \\
\hline
\end{tabular}
\label{tab: mtptab}
\end{table}
Tab.~\ref{tab: mtptab} compares the proposed method with a previous work GRIP~\cite{li2019grip} on the MTP task. It shows that the proposed model, when applied to multi-vehicular trajectory prediction, matches the previous work in terms of RMSE.

\begin{figure}
    \centering
    \includegraphics[trim={0cm 0cm 0cm 0cm}, clip, width=1.0\columnwidth]{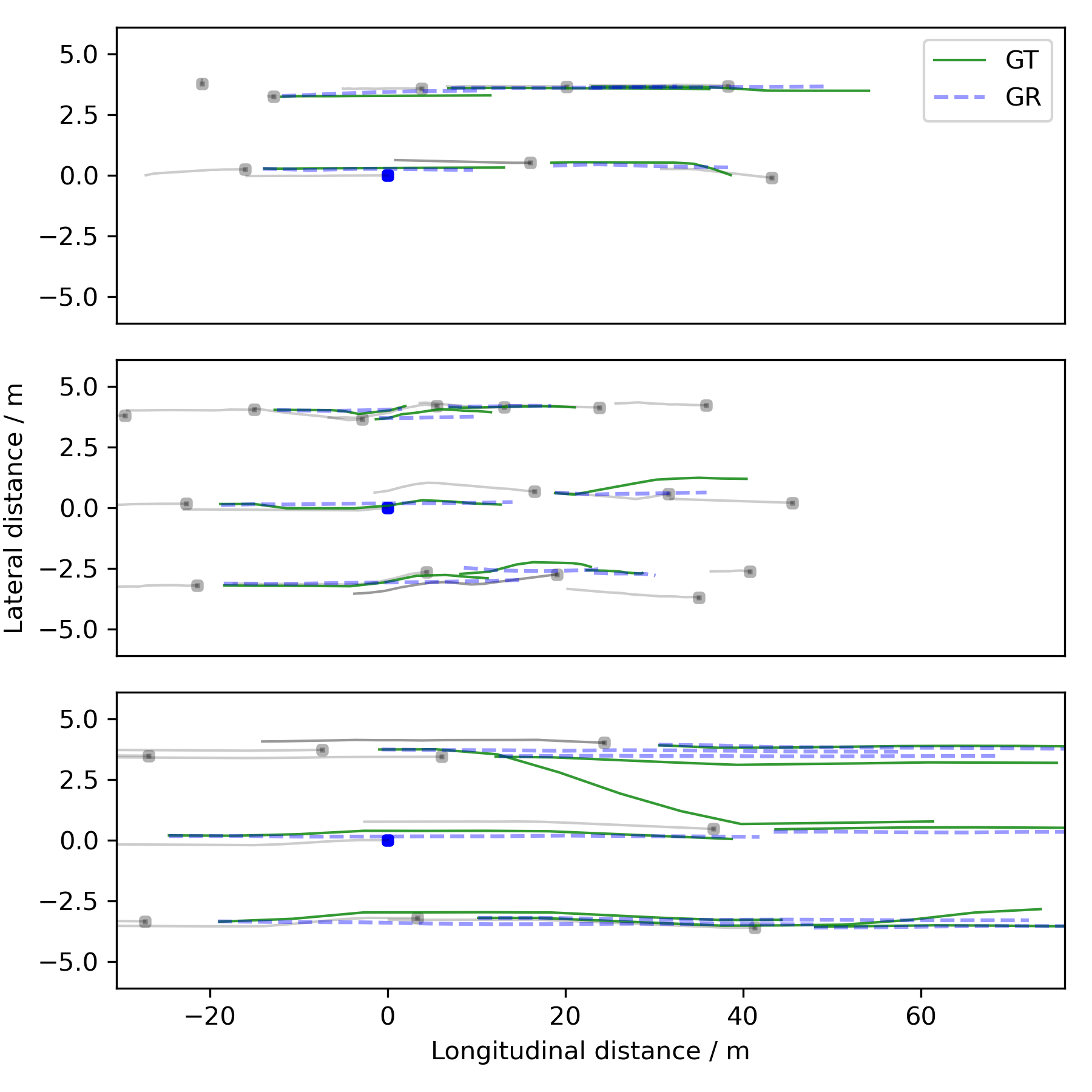}
    \vspace{-7mm}
    \caption{\textbf{Visualized MTP predictions.} Blue square is the ego vehicle and gray squares represent the rest of considered vehicles. Only future trajectories of target vehicles are plotted. Green lines are the ground truth and dashed blue lines are the prediction future trajectory. All the vehicles move from left to right.}
    \label{fig: mtpvis}
    \vspace{-5mm}
\end{figure}
Fig.~\ref{fig: mtpvis} visualizes the prediction results of the proposed model on the MTP task. It can be seen that the proposed method can predict the multiple trajectories longitudinally while it fails to predict the lane-change maneuver in the next 5 seconds. This can be explained by the imbalance of the MTP dataset since the majority of the future trajectories in the dataset are keeping lane, and it is hard to get a roughly balanced dataset for MTP.

\section{CONCLUSIONS}
\label{sec: conclusion}
This work proposes a GNN-RNN-based method for trajectory prediction to model the inter-vehicular interaction among various vehicles. RNN is used to capture the dynamics feature of vehicles, and GNN is adopted to summarize the interaction feature. Another RNN serves as the decoder jointly considers the dynamics and interaction feature for prediction. This work finds that both the target vehicle's individual dynamics feature and its interaction with other vehicles affect the prediction accuracy. The proposed method matches state-of-the-art methods on the NGSIM dataset in terms of RMSE. 

This work can be improved to handle multi-vehicular trajectory prediction properly, which is necessary for the downstream decision-making module of autonomous driving. It can also be extended to consider the multi-modality of driving behaviors. 

\addtolength{\textheight}{-2.5cm}   
\section*{ACKNOWLEDGMENT}
This work was supported in part by A*STAR Grant (No. 1922500046), Singapore, the Alibaba Group through Alibaba Innovative Research (AIR) Program and Alibaba-NTU Singapore Joint Research Institute (JRI) (No. AN-GC-2020-012), A*STAR AME Young Individual Research Grant (No. A2084c0156), and the SUG-NAP Grant, Nanyang Technological University, Singapore.

\bibliographystyle{unsrt}
\bibliography{gr.bib}
\end{document}